# Improving Hand Recognition in Uncontrolled and Uncooperative Environments using Multiple Spatial Transformers and Loss Functions

[1]Wojciech Michal Matkowski, [2]Xiaojie Li and Adams Wai Kin Kong

*Abstract*—The prevalence of smartphone and consumer camera has led to more evidence in the form of digital images, which are mostly taken in uncontrolled and uncooperative environments. In these images, criminals likely hide or cover their faces while their hands are observable in some cases, creating a challenging use case for forensic investigation. Many existing hand-based recognition methods perform well for hand images collected in controlled environments with user cooperation. However, their performance deteriorates significantly in uncontrolled and uncooperative environments. A recent work has exposed the potential of hand recognition in these environments. However, only the palmar regions were considered, and the recognition performance is still far from satisfactory. To improve the recognition accuracy, an algorithm integrating a multi-spatial transformer network (MSTN) and multiple loss functions is proposed to fully utilize information in full hand images. MSTN is firstly employed to localize the palms and fingers and estimate the alignment parameters. Then, the aligned images are further fed into pretrained convolutional neural networks, where features are extracted. Finally, a training scheme with multiple loss functions is used to train the network end-to-end. To demonstrate the effectiveness of the proposed algorithm, the trained model is evaluated on NTU-PI-v1 database and six benchmark databases from different domains. Experimental results show that the proposed algorithm performs significantly better than the existing methods in these uncontrolled and uncooperative environments and has good generalization capabilities to samples from different domains.

## I. Introduction

Biometric characteristics such as face, DNA, fingerprint and tattoo are commonly used as evidence or clues by law enforcement agencies in forensic investigation. Latent fingerprints and DNA need to be physically collected at a crime scene, which may be unknown. Furthermore, they may be hard to collect, e.g., due to large crowds in public places during riots or looting. Currently, with the prevalence of cameras, more evidence is in the form of digital multimedia, such as images. In digital and multimedia forensics, identification of terrorists, rioters, child sexual offenders and their victims is challenging if no obvious traits such as face or tattoos are well-visible.

Terrorists commonly cover their faces using masks, e.g., in terrorist propaganda images. Child sexual offenders hide their identities by trying not to expose obvious traits such as face or tattoo in child sexual abuse materials (CSAM). Rioters, captured by press photographers also use masks or clothes to cover their faces. However, other body parts, e.g., hands might be exposed [1, 2] in images and can be used for identification. Fig. 1 shows examples of images with exposed hands that might be use for forensic investigation. Such images can be collected from the Internet, social media accounts, directly from the press, subject's families, suspect's hard drives, citizen's tips, police bookings or photo releases by law enforcement. For example, after the recent unrest in the US Capitol Building on the 6[th] January 2021, Federal Bureau of Investigation (FBI) and Metropolitan Police Department released images (see Fig. 1, the first row) of persons of interest to identify and are seeking for more information including images and videos that could be relevant to investigation.

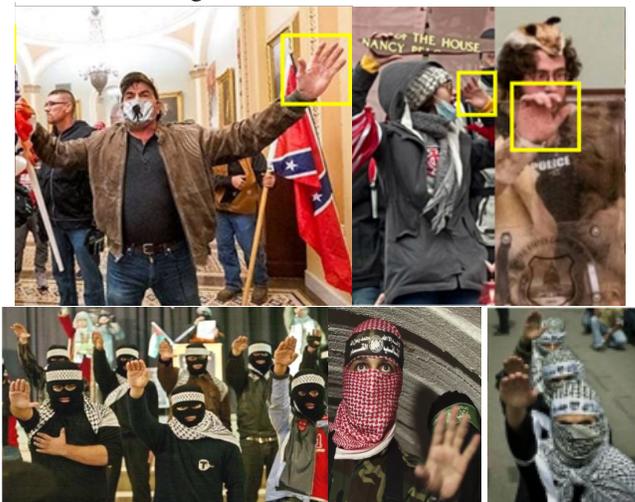

Fig. 1. The first row is examples images of persons of interest in unrest-related offenses released by law enforcement after unlawful entry to the US Capitol. Hands are highlighted in yellow boxes. The second row is terrorist images. Note that CSAM also contains hands in images. However, they cannot be shown here due to legal reasons.

Manuscript received [date]. This work is partially supported in part by the NTU Internal Funding - Accelerating Creativity and Excellence (NTU–ACE2020-03).

W.M. Matkowski is with Thales, 12 Ayer Rajah Crescent, Singapore 139941. E-mail: wojciech.matkowski@ntu.edu.sg

X. Li is with College of Information Science and Engineering, Shandong Normal University, Jinan, China E-mail: xli162011@163.com.

A.W.K. Kong is with the School of Computer Science and Engineering, Nanyang Technological University, Singapore, 639798. E-mail: adamskong@ntu.edu.sg.

[1]This work was done when W.M. Matkowski worked for the Nanyang Technological University.

[2] A.W.K. Kong is the corresponding author.

In the biometric community, hand-based biometric traits have been studied for many years. Many recognition methods based on various biometric traits, e.g. hand geometry [3-5], palmprint [6-8], finger vein [9], fingerprint [10-12], finger [13], 2D/3D finger knuckle patterns [14-17], thermal information [18], and multimodal features [19] have been proposed for commercial and forensic applications. Except for latent fingerprints and latent palmprints recognition methods, almost all of these methods are designed for commercial applications, such as access control, where images are collected from special-designed data acquisition systems with user cooperation. For example, 3D finger knuckle image collection requires a fixed camera, seven evenly distributed illuminations and a control system [15]. Therefore, they cannot be applied directly to the hand images collected in the aforementioned cases. As shown in Fig. 1, images collected in these real-world scenarios have very different visual attributes from the images collected in laboratory-controlled environments or special-designed data acquisition systems, due to the presence of cluttered background, partial occlusions, diverse image resolutions, poses, and illumination, etc. Thus, hand recognition based on images from these real-world scenarios for forensic investigation is more challenging.

Compared to the methods designed for commercial applications, hand recognition using images from the uncontrolled and uncooperative environments mentioned before has received very little attention. To tackle this problem, Matkowski et al. proposed a palmprint recognition method based on images from uncontrolled and uncooperative environments for forensic investigation [20]. Experimental results showed that hand images collected from these real-world scenarios can provide biometric clues for personal identification. However, their method only focuses on the palmar region and identity information in full hand images is not fully exploited. Thus, more research efforts are desirable to tackle this problem.

In this paper, to study hand recognition for forensic investigation, over 250,000 hand key points are provided for a publicly available database (which contains 2035 palmar hands), namely NTU-PI-v1 and a deep learning-based algorithm is proposed. The proposed end-to-end algorithm operates on full hand images and consists of a training scheme using a combination of multiple softmax and triplet loss functions, a Multi-Spatial Transformer Network (MSTN) to localize and align palm, fingers and thumb and multiple feature extraction backbones and embedding modules to extract their corresponding feature embedding. To the best of our knowledge, the proposed hand recognition algorithm is the first work to use full hand images collected from uncontrolled and uncooperative environments for forensic investigation.

The proposed algorithm is evaluated on NTU-PI-v1 and palm, fingers and thumb contributions to accuracy are reported. In identification experiments, the proposed algorithm outperforms the second-best result from the literature by a margin of 20.5%. In addition, the proposed algorithm is compared with relevant state-of-the-art methods demonstrating its good generalization capabilities (without any fine-tuning) across six publicly available palmprint databases from relevant but different domains.

The main contributions of this work are five-fold.

1) According to our best knowledge, it is the first end-to-end deep algorithm operating on full hand images for uncontrolled and uncooperative environments. To incorporate the fingers into recognition, 256,773 finger key points are marked on the images in NTU-PI-v1.

2) The proposed algorithm achieves an improvement over the second-best on NTU-PI-v1 by up to 20.5%, attributing to the proposed training polices, multi-spatial transformer network (MSTN), the design of embedding modules and the incorporation of more information from finger regions.

3) The proposed MSTN uses a single network to regress hand and finger parameters and six grid generators and samplers for palm and finger segmentation and alignment. This design can achieve better estimation accuracy and higher quality of alignment for recognition.

4) The proposed network has multiple subnetworks for extracting features from fingers and palm and aligning them. To achieve high quality fingers and palm features and enhance alignment performance, an objective function, which is composed of multiple loss functions for fingers, palm and MSTN, is proposed to improve overall performance.

5) The model trained on NTU-PI-v1 is applied on six different public palmprint datasets without fine-tuning and the experimental results show that the model has good generalization capabilities to samples from different domains, which demonstrates that NTU-PI-v1 enables a more robust model training.

The rest of this paper is organized as follows. The related palmprint and finger recognition methods are introduced in Section II. In Section III, the proposed hand recognition algorithm is described. The experimental results and analyses on benchmark databases are presented in Section IV. Finally, Section V gives some concluding remarks.

## II. Related Work

The algorithm proposed in this paper aims to extract hand features from images collected from unconstrained environments for personal recognition. Thus, related works, including palmprint recognition methods and finger recognition methods are reviewed.

Palmprint is deemed to be a hard biometric trait since it contains many discriminative and permanent features, i.e. flexion creases, wrinkles, ridges, and minutiae [21]. Palmprint recognition based on images has received great research attention in the past two decades. Generally, palmprint images can be classified into four categories in terms of the acquisition process, i.e. the contact-based, contactless, latent and 3-D palmprint images [22]. Among them, contactless palmprint images, referring to images collected without any contact between the hand and the surface of the acquisition devices, are the closest ones to the cases discussed in this paper.

Typically, a traditional palmprint recognition framework mainly contains the following procedures: image acquisition, preprocessing, feature extraction and matching [21]. In the past two decades, both conventional hand-crafted feature-based and deep learning-based methods have been proposed for palmprint recognition and achieved remarkable performance on the public databases.

Line orientation and texture are critical clues to palmprint recognition; thus, both generic and palmprint-specific descriptors were designed to extract orientation and texture information. Line features were extracted explicitly using line or edge detectors for palmprint recognition [23-26]. Recently, Palma et al. proposed a recursive and dynamic algorithm for flexion creases matching [27]. In the matching procedure, a positive linear dynamical system was employed and evolved based on the matching level of the two input palm images. The proposed method could reduce the impact of noise effectively and thus was robust in dealing with low-resolution and noisy images. However, since people may have similar flexion creases and wrinkles are too sensitive to noise and illumination changes, accurately extracting them for recognition is difficult [22].

Coding-based methods, such as PalmCode [28], binary orientation co-occurrence vector (BOCV) [7], CompCode [6], OrdinalCode [29], and Double Orientation Code (DOC) [30] were proposed to extract orientation information by applying a sets of filters and generate a bitwise representation for high speed matching. Coding-based methods achieve good performance on contact-based palmprint recognition. But, the misalignment occurred in the contactless settings deteriorates their performance [22]. Thus, some researchers employed local line and texture descriptors, which are more robust to local variations in rotation, scale, translation and illumination. For example, Scale Invariant Feature Transform (SIFT), Local Binary Pattern (LBP) [31, 32], Local Directional Patterns (LDP) [33], CR-CompCode [34], local micro-structure tetra pattern [35], local line directional pattern (LLDP) [36] and histogram of oriented lines (HOL) [37] descriptors were utilized for palmprint recognition. In addition, a learning-based method was proposed to extract discriminant direction binary codes (DDBCs) for each pixel of palmprint image and concatenate them to form the discriminant direction binary palmprint descriptor [38]. These conventional palmprint descriptors required hand crafted filters, which were designed based on the empirical experience. Consequently, the performance of these methods highly depends on the selected filters and the corresponding parameters.

The success of deep neural network has enabled promising deep learning methods for palmprint recognition, in which Convolutional Neural Networks (CNNs) are incorporated for better feature extraction and representation. For contactless palmprint recognition, Fei et al. tested four classic deep learning architectures, i.e. AlexNet [39], VGG-16 [40], GoogLeNet [41] and ResNet-50 [42]. All of them were pretrained using the ImageNet dataset [43] and fine-tuned on palmprint images. Experimental results showed that these architectures achieved comparable or even better performance than conventional palmprint recognition methods [22]. Minaee and Wang proposed a three-stage palmprint recognition method using deep network as feature extractor [44]. Scattering features were extracted using a pretrained deep scattering convolutional network (DSCN) [45] and further processed using principal component analysis (PCA) for dimensional reduction. Support vector machine (SVM) and minimum-distance classifier were employed as classifiers. The method was examined on contact-based datasets. Similarly, in [46], pretrained AlexNet was applied and combined with SVM to perform palmprint recognition for newborns. The evaluations on contactless palmprints demonstrated that the method outperformed coding-based methods. Genovese et al. designed PalmNet, which combined Gabor filters and CNN, to extract palmprint-specific descriptors for contactless palmprint recognition in an unsupervised manner [47]. Zhao and Zhang proposed deep discriminative representation (DDR) to extract high-level discriminative features for palmprint recognition in different scenarios [8]. To improve the performance of coding-based methods, which require a set of predefined orientation filters, Zhao et al. proposed to learn the complete and discriminative direction pattern, which contains both the local direction feature and underlying structural information, for palmprint recognition [48].

A major disadvantage of deep learning based palmprint recognition methods is the requirement of a large amount of annotated data for model training. To deal with this problem, Shao et al. proposed a meta-Siamese network (MSN), which integrated meta-learning strategy into Siamese network, for few-shot palmprint recognition [49]. In addition to mean square error based similarity loss, contrastive loss and binomial deviance were also employed to maximize the feature distance between different categories of samples while reducing the distance between samples from the same category. Similarly, Svoboda et al. employed a Siamese network based on the AlexNet [39] model for palmprint recognition [50]. A d-prime index related loss function describing the separation between the genuine and impostor distributions was utilized to optimize the network. The method was tested on contactless palmprint databases and achieved good performance. Considering the undersampling problem in palmprint recognition, i.e. insufficient training samples for each individual, Zhao and Zhang presented a joint constrained least-square regression (JCLSR) framework [51]. With this method, different local regions were regularly sampled from the palmprint image and deep local convolution features with positional information were learned from these subregions to generate a complete and discriminative description of palmprint image. A regularization constraint was imposed to the least-square regression (LSR) model to guarantee that the learned projected matrices of the subregions from the same palmprint image were similar. A deep learning-based framework for contactless palmprint detection and identification is proposed in [52]. A soft-shifted triplet loss function was designed to learn discriminative palmprint features and Faster R-CNN architecture was adapted to detect palmprint region in an automated manner. The proposed method was examined on different databases and experimental results show that the method has good performance.

Although palmprint recognition methods based on deep neural network have given promising results, the performance of these methods may degrade when they are applied to samples from different domains, e.g., images collected under different environments or using different devices. To address this problem, Shao et al. proposed a joint pixel and feature alignment (JPFA) framework, in which two alignment models, i.e. deep style transfer model and deep domain adaptation model, were adopted to extract domain-aligned features in source and target datasets, for cross-dataset palmprint recognition [53]. Du et al. proposed a regularized adversarial domain adaptative hashing (R-ADAH) framework for cross-domain palmprint recognition [54].

In some contactless palmprint recognition, the acquisition devices and environments used in training data collection may be different from those used in testing data collection, e.g., mobile phone based palmprint recognition. Thus, methods and databases were developed for this type of contactless palmprint recognition. Based on the knowledge distillation mechanism, Shao et al. proposed a deep distillation hashing (DDH) algorithm, in which VGG-16-based deep hashing network was employed as a teacher network and a light model containing only two convolutional layers and three fully connected (FC) layers was constructed as the student network, for efficient deep palmprint recognition by converting the palmprint images into binary code [55]. An adversarial metric learning method, which took the distribution of palmprint in the holistic feature space into consideration, was proposed to improve the palmprint recognition accuracy [56]. In addition, a contactless palmprint database named XJTU-UP, in which images were collected by five different devices, was established. Although the image collection setting, e.g., lighting, background, was not strictly controlled, volunteers were asked to take images in a cooperative manner. Thus, images from XJTU-UP are very different from the images in forensic applications mentioned before.

All the pervious methods were designed for commercial applications, where close-up palmprint images were collected with great user cooperation. Thus, palmprint images used in the previous studies rarely suffered from motion blur, serious occlusion, extreme low-resolution, etc. However, these are common problems for palmprint images collected for forensic applications. Furthermore, they still relied on traditional image processing approaches [28] for ROI extraction. Consequently, even though deep learning was used, they are not fully end-to-end. Matkowski et al. established a new palmprint database named NTU-PI-v1, in which images were collected from Internet and taken in uncontrolled and uncooperative environment [20]. A deep learning framework was designed to perform ROI extraction, alignment, feature extraction and matching in an end-to-end manner. Matkowski et al.'s work only used palmar region for recognition but ignored information in fingers and thumb. Using fingers and thumb for personal identification in these uncooperative and uncontrolled environments is challenging, because of their great flexibility of movement.

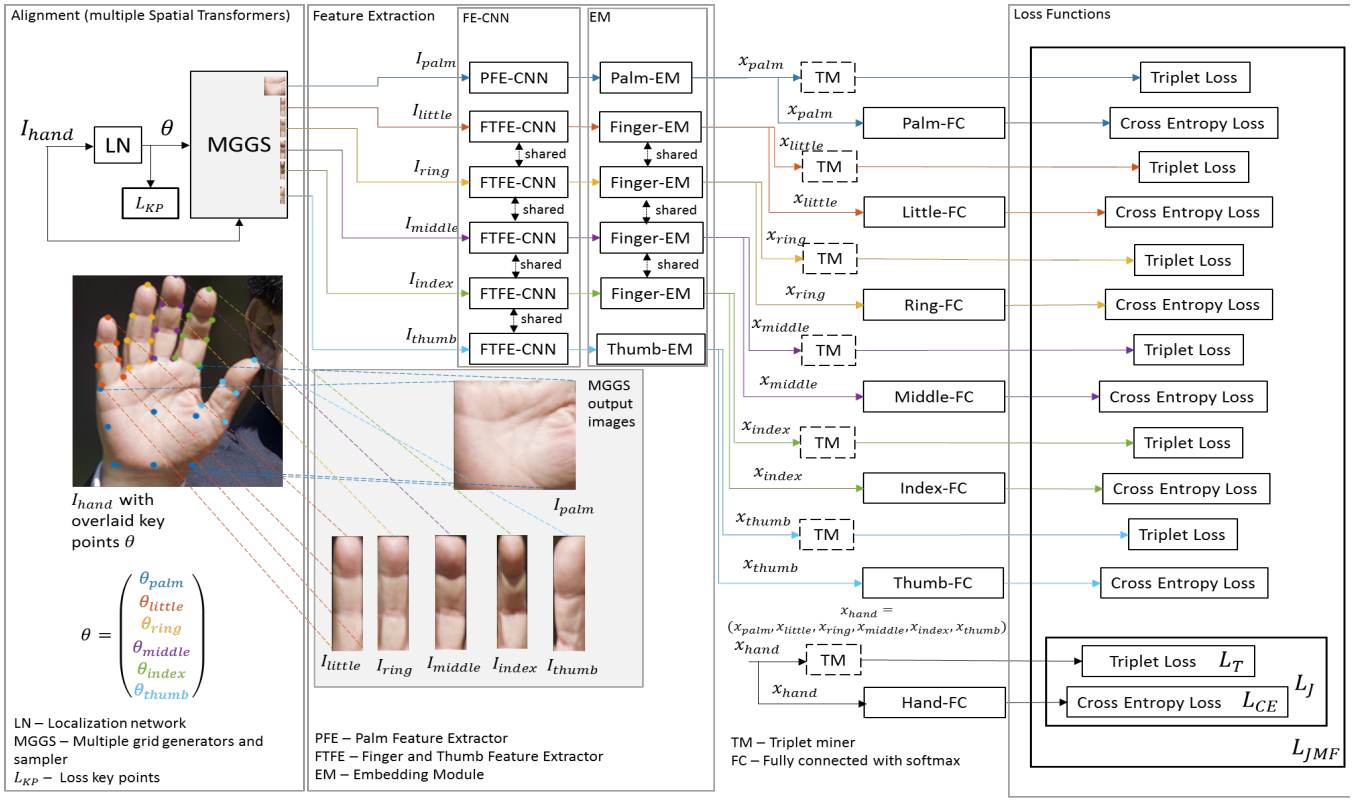

Fig. 2. The proposed algorithm architecture containing the proposed multiple Spatial Transformer Network (MSTN) and loss functions used for training.

Comparing to palmprint recognition, relatively less attention has been paid to personal recognition based on finger images, including finger [13] and 2D /3Dfinger knuckle patterns [14-17]. These methods require high resolution images, whereas it is not applicable for unconstrained environments without subject's cooperation. Thus, they are not suitable to the case discussed in this paper.

## III. THE PROPOSED ALGORITHM

The proposed algorithm is a multiple spatial transformer network (MSTN), that contains multiple Spatial Transformers, feature extractors, embedding modules and a training scheme with multiple loss functions. The proposed algorithm uses multiple Spatial Transformers for hand image alignment (Section III-A) and Feature Extractor CNNs (FE-CNN) and Embedding Modules (EM) for hand feature extraction (Section III-B). The training scheme employs L2 and L1 loss and multiple fully connected layers (FC) with cross entropy loss and triplet loss functions (See Section III-C). Fig. 2 illustrates the proposed algorithm.

### A. Alignment

In this paper, Spatial Transformers [57] which enable image spatial transformations are used to align palm, four fingers and thumb images. Previous works utilized spatial transformer network to hypothesize alignment parameters for palm [20] and fingerprint alignment [58], [59], in which spatial transformer network was incorporated into a standard neural network architecture. A Spatial Transformer contains a Localization network (LN), which is a pretrained backbone CNN, and grid generator and sampler (GGS). LN takes full hand image $I_{hand}$ as an input and directly regresses the hand key points' coordinates $\theta = (\theta_{palm}, \theta_{little}, \theta_{ring}, \theta_{middle}, \theta_{index}, \theta_{thumb})$. In the previous method [20], for end-to-end palmprint recognition, Spatial Transformer with $\theta_{palm}$ focused on palmar regions. Thus, it contained a single GGS. In this work, to handle full hand images, multiple GGS (MGGS) are used. MGGS generates six deformation grids $T_{\theta_m}$ based on $\theta_m$ using thin plate spline (TPS) [60]. Deformation grids $T_{\theta_m}$ are used to separately align palm, fingers and thumb to their corresponding template grids $G_m$, where $m \in M$ and $M = \{palm, little, ring, middle, index, thumb\}$. For each $m$, MGGS performs alignment by bilinear sampling of $I_{hand}$ based on $T_{\theta_m}$ into $G_m$ and returns six images $I_{palm}, I_{little}, I_{ring}, I_{middle}, I_{index}, I_{thumb}$, which are ROIs of palm, little finger, ring finger, middle finger, index finger and thumb, respectively.

### B. Feature Extraction

In the proposed algorithm, features are extracted from six different images: 1 palm, 4 fingers and 1 thumb. Therefore, six sub-networks are used. Each sub-network has the same structure containing a feature extractor, which is a pretrained backbone CNN (FE-CNN) and an embedding module (EM) with newly initialized weights. The palm image $I_{palm}$ is input to Palm Feature Extractor CNN (PFE-CNN) followed by Palm Embedding Module (Palm-EM), which returns palm features $x_{palm}$. All the finger images $I_{little}, I_{ring}, I_{middle}, I_{index}$ are input to four Finger and Thumb Feature Extractor CNNs (FTFE-CNN) followed by four Finger Embedding Modules (Finger-EM), which return the corresponding finger features $x_{little}$, $x_{ring}$, $x_{middle}$, $x_{index}$. The thumb image $I_{thumb}$ is input to FTFE-CNN followed by Thumb Embedding Module (Thumb-EM) returning thumb features $x_{thumb}$. In these sub-networks, the learnable parameters are shared among all FTFE-CNNs and among all the Finger-EMs because they operate on finger images with similar texture and crease features. The final hand feature is obtained by concatenating all the features i.e., $x_{hand} = (x_{palm}, x_{little}, x_{ring}, x_{middle}, x_{index}, x_{thumb})$ and is passed to a fully connected layer with softmax.

### C. Loss Functions

Typically, in deep learning based biometric recognition, cross entropy or/and triplet loss functions are used during training.

Triplet loss can be defined as follows:

$$L_T(X^a, X^p, X^n) = \sum_{(x^a, x^p, x^n) \in B} \max[(d(x^a, x^p) - d(x^a, x^n) + \alpha), 0], \quad (1)$$

where $d$ is a distance function, e.g., Euclidean distance, $\alpha > 0$ is a margin parameter and $B$ is a set of selected triplets $(x^a, x^p, x^n)$, within a batch containing anchor, positive and negative samples, respectively. Cross-entropy loss can be defined as follows:

$$L_{CE}(Y, P) = -\sum_{i=1}^{N} \sum_{c=1}^{C} y_{ic} \cdot \log(p_{ic}), \quad (2)$$

where $y_{ic}$ is 1 if the $i^{th}$ sample is from class $c$; otherwise $y_{ic}$ is 0, $p_{ic}$ is the softmax output probability of class $c$ for the $i^{th}$ sample, $N$ is the batch size and $C$ is the number of classes. In the embedding space, triplet loss "pushes" away samples $x^a, x^n$ from different class and "pulls" samples $x^a, x^p$ from the same class closer to each other, which promotes higher within-class compactness. However, the number of all possible triplets grows cubically with respect to number of training samples making the convergence of triplet loss relatively slow and heavily dependent on triplet sampling. Cross-entropy loss, which promotes separation between different classes, converges much faster than triplet loss but does not directly encourage within-class compactness, which may result in poor regularization of the feature embedding layer and less discriminatory features. Therefore, joint loss $L_J = L_T + L_{CE}$ is likely to provide benefits in terms of feature embedding's quality and higher accuracy.

Extracted from the sub-networks, hand features $x_{hand}$ (see Section III-B), contain concatenated features from six palm, fingers and thumb images. However, directly optimizing $x_{hand}$ does not ensure that each feature in $x_m$ will contribute to recognition accuracy after training. For example, features $x_{palm}$ maybe much easier to learn than the others. Therefore, it would overtake the focus on palm, neglecting the contributions of fingers and thumb. To ensure balanced focus and contribution of $x_{hand}$ and all other features $x_m$, a joint multi-feature loss is proposed. The final joint loss function $L_{JMF}$ over a set of feature types $M'$ can be expressed as follows:

$$L_{JMF} = \sum_{m \in M'} \lambda_{CE_m} L_{CE}(Y_m, P_m) + \sum_{m \in M'} \lambda_{t_m} L_T(X_m^a, X_m^p, X_m^n) + \lambda_{KP} L_{KP}, \quad (3)$$

where $\lambda s$ are loss weight parameters, $M' \subseteq (M \cup \{hand\} = \{palm, little, ring, middle, index, thumb, hand\})$, and $L_{KP}$ is a loss for the Localization network. The Localization network is supervised by L2 loss: $L_{L2}(\Theta_{GT}, \Theta) = \frac{1}{N}\sum_{i=1}^{N}(\theta_{GT} - \theta)^2$ and L1 loss: $L_{L1}(\Theta_{GT}, \Theta) = \frac{1}{N}\sum_{i=1}^{N}|\theta_{GT} - \theta|$, where $\theta_{GT}$ are the ground truth coordinates of the hand key points (see Section IV-A) and $N$ is the number of samples in a batch. The final loss for Localization network, which estimates hand key points is defined as follows:

$$L_{KP} = \lambda_{l2} L_{L2}(\Theta_{GT}, \Theta) + \lambda_{l1} L_{L1}(\Theta_{GT}, \Theta). \quad (4)$$

### D. Implementation Details

The proposed algorithm is implemented in Python using PyTorch library and its pretrained backbone models. For the Localization network LN of the Spatial Transformer (ST), ResNet-50 is used, with the last layer replaced by newly initialized fully connected layer with 84 neurons (42 key points defined by 2 coordinates) which regresses hand key points $\theta$. The full hand images $I_{hand}$ are resized into 128 x 128 pixels, normalized according to network's requirements and input to LN. LN is initially fine-tuned for 100 epochs using $L_{KP}$ loss, with ADAM optimizer, 0.00001 learning rate and batch size 16. During training, for data augmentation, color jittering and random combinations of random rotations ([0, 360] deg.), translations ([-0.25, 0.25]), and scaling ([0.8, 1.2]) are applied on the fly to full hand images and the corresponding ground truth key points $\Theta_{GT}$. In the ST, multi grid generators and samplers take input full hand image $I_{hand}$ resized to 227 x 227 pixels and align, palm, fingers and thumb based on $\theta$ to their corresponding templates.

In the feature extraction, the aligned images' sizes are 128 x 128 pixels for palm and 128 x 32 pixels for each finger and thumb. All the FE-CNNs, namely PFE-CNN and FTFE-CNNs are ResNet-50 pruned after the 6[th] block. Therefore, the dimensions of FE-CNN outputs are 16 x 16 x 512 for palm and 16 x 4 x 512 for fingers and thumb. The relatively high spatial size is to preserve spatial structure of the features from the aligned images. For the Embedding Module EM, two variants with newly initialized parameters are proposed: 1) EM-conv - convolutional layer followed by Tanh activation and batch normalization and 2) EM-fc - fully connected layer followed by batch normalization. Note that the batch normalization in both EMs is essential and without it the accuracy drops. In the EM-fc, the number of output neurons is set to 512 for palm's Palm-EM and 128 for each fingers' Finger-EM and 128 for thumb's Thumb-EM. In each EM-conv, convolutional layer is set to have 16 filters of 1 x 1 x 512 size. Thus, EM-conv preservers the spatial size of the preceding layer of a FE-CNN, and outputs 16 spatial feature maps.

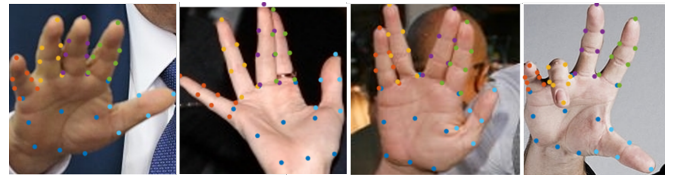

Fig. 3. Examples of images NTU-PI-v1 database with marked hand key points.

After LN has been initially fine-tuned, its layers are frozen and all the modules are connected to form the proposed algorithm which is trained for 200 epochs, with 256 batch size using the proposed $L_{JMF}$ loss. FE-CNN and EM are optimized by ADAM with learning rates set up to 0.0001 and 0.001, respectively. In the triplet miner (TM), the triplets within a batch are selected on the fly using a batch hard strategy [61]. Note that the proposed algorithm is end-to-end trainable and the gradients from $L_{JMF}$ are also propagated to the LN. After 20 epochs, LN's last layer is unfrozen with learning rate 0.00001. To control the stability of the spatial transformer in the end-to-end training, $L_{KP}$ is used after 20 epochs to regularize LN. In the loss functions, the hyperparameters $\lambda_{l1}, \lambda_{l2} \lambda_{t_m}$ are set to 1, 0.1, 40, respectively and $\lambda_{CE_m}$ is decayed every epoch $\lambda_{CE_m} = 1/epoch$. In all the experiments, 1 nearest neighbor (1-NN) using cosine similarity on extracted features is used for recognition. The training is run on a single Tesla V100 32GB GPU card and takes slightly more than 1 hour.

## IV. EXPERIMENTS

### A. Databases and Protocols

The proposed algorithm is mainly evaluated on NTU-PI-v1 database [20] because it is the only publicly available database containing hand images from uncontrolled and uncooperative environments. This database aims to simulate diversities in forensic scenarios, where hand images are taken in uncontrolled environment without user cooperation and goal of personal recognition. The diversity in NTU-PI-v1 is reflected in hand pose, background, lightening, image resolution, and hand appearance related to subjects' sex, ethnicity and age. NTU-PI-v1 contains 7881 images from 2035 different hands and provides 70,929 key points, which can be used to extract and align palmprint region of interest (ROI). The median size of NTU-PI-v1 full hand images is only 115x115 pixels, which is significantly smaller than other contactless palmprint images. Examples of NTU-PI-v1 images can be found in Fig. 4(g). More detailed comparison between NTU-PI-v1 and other contactless palmprint databases can be found in [20]. So far, NTU-PI-v1 is the only publicly available hand image database from such environment, which is a reasonable approximation for forensic images, such as child sexual abuse materials (CSAM), terrorist or rioter images, where hand may appear.

In this paper, additional new hand key points are provided for all fingers and thumb. Fig. 3. shows examples of hand key points on images from NTU-PI-v1. There are 7 and 5 key points manually marked for each finger and thumb, respectively. Totally, 256,773 new hand key points are provided and used for training as $\theta_{GT}$. In the experiments, the database split and evaluation protocol for NTU-PI-v1 are the same as in [20]. Cumulative match characteristics (CMC) curve, which is a common performance index in forensic scenarios such as search in a database, is used as evaluation metric. Moreover, rank-1 and rank-30 identification accuracies are reported. Note that higher ranks such as rank-30 are also useful in forensics because they narrow down the list of subjects to search by forensic investigator.

In addition, six related and representative contactless palmprint databases are used in evaluation, namely NTU-CP-v1 [20], IITD Touchless [62], PolyU Contactless 2D [63], CASIA-Palmprint [64], REST [65] and NUI_Palm1 [66]. In the experiments, for the CASIA, PolyU, IITD, REST and NTU-CP-v1, database split into gallery and probe sets is the same as in [20]. For NUI_Palm1, first four images from one smartphone device tagged as "G4" are put into the gallery set and the remaining ones, which are from another four different devices are put into the probe set. Most of these databases, in fact, contain hand images with visible fingers which are usually not utilized for recognition with few with exceptions in [67], [68]. Note that the images from these databases except for NTU-PI-v1 come from relatively well-controlled and cooperative environment.

Because the proposed algorithm requires square-sized full hand images, simple automatic image processing techniques are applied to obtain square images from CASIA, PolyU, IITD, and REST, preserving the aspect ratio. For CASIA, images are padded with zeros on left and right sides. For IITD, fixed central part of the image is cropped. For PolyU and REST, Otsu thresholding and the square bounding box around the biggest connected component are used. In NUI_Palm1, images have relatively diverse background and thus, palmprint key points which are provided in this database are used to crop full hand image and pad with zeros if the crop is outside of the boarder. Fig. 4 shows examples of square-sized images from these six contactless palmprint databases and examples of full hand images from NTU-PI-v1. Note that in the experiments, the left hand images in all the databases are flipped into the right and considered as different subject, which is a common practice to increase the number of possible comparisons.

TABLE I
RANK-1 AND RANK-30 ACCURACIES (%) OF DIFFERENT FEATURES USING EM-FC AND DIFFERENT LOSS FUNCTIONS ON NTU-PI-V1.

|  | palm | | little | | ring | | middle | | index | | thumb | | hand | |
| --- | --- | --- | --- | --- | --- | --- | --- | --- | --- | --- | --- | --- | --- | --- |
|  | Rank-1 | Rank-30 | Rank-1 | Rank-30 | Rank-1 | Rank-30 | Rank-1 | Rank-30 | Rank-1 | Rank-30 | Rank-1 | Rank-30 | Rank-1 | Rank-30 |
| $L_{CE}$ | 28.96 | 54.64 | 7.72 | 26.30 | 11.03 | 32.66 | 10.53 | 33.28 | 9.70 | 30.38 | 7.07 | 25.08 | 34.79 | 63.31 |
| $L_T$ | 43.22 | 66.83 | 4.43 | 23.31 | 6.68 | 28.57 | 6.63 | 30.65 | 6.12 | 28.19 | 4.76 | 23.13 | 43.57 | 67.48 |
| $L_J$ | 43.78 | 67.18 | 3.87 | 22.60 | 7.15 | 27.89 | 5.85 | 30.02 | 5.86 | 27.78 | 4.67 | 22.81 | 44.57 | 67.57 |
| $L_{JMF}$ | **45.02** | **67.98** | **17.66** | **42.54** | **23.57** | **50.97** | **23.10** | **51.39** | **21.71** | **47.60** | **18.10** | **40.71** | **56.62** | **80.35** |

TABLE II
RANK-1 AND RANK-30 ACCURACIES (%) OF DIFFERENT FEATURES USING EM-CONV AND DIFFERENT LOSS FUNCTIONS ON NTU-PI-V1.

|  | palm | | little | | ring | | middle | | index | | thumb | | hand | |
| --- | --- | --- | --- | --- | --- | --- | --- | --- | --- | --- | --- | --- | --- | --- |
|  | Rank-1 | Rank-30 | Rank-1 | Rank-30 | Rank-1 | Rank-30 | Rank-1 | Rank-30 | Rank-1 | Rank-30 | Rank-1 | Rank-30 | Rank-1 | Rank-30 |
| $L_{CE}$ | 33.99 | 59.98 | 9.37 | 27.21 | 14.67 | 36.21 | 13.90 | 36.33 | 12.63 | 32.13 | 10.14 | 28.37 | 38.63 | 62.39 |
| $L_T$ | 49.37 | 73.52 | 10.71 | 33.87 | 15.35 | 41.15 | 16.44 | 43.28 | 13.66 | 38.19 | 8.90 | 29.55 | 60.91 | 83.90 |
| $L_J$ | **52.42** | 75.29 | 10.20 | 32.45 | 12.84 | 36.39 | 16.21 | 41.36 | 12.75 | 36.39 | 9.97 | 30.94 | 61.68 | 84.43 |
| $L_{JMF}$ | 51.44 | **75.59** | **15.46** | **40.05** | **21.71** | **47.84** | **22.01** | **48.07** | **18.63** | **44.31** | **15.35** | **38.16** | **62.42** | **84.76** |

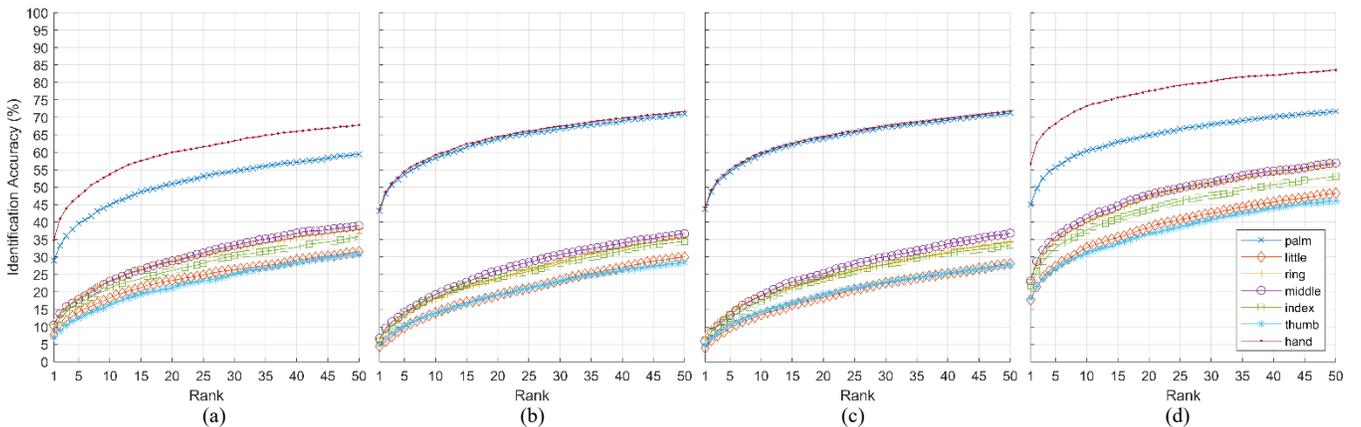

Fig. 5. CMC curves of different features extracted from palm, little, ring, middle, index fingers, thumb and hand, after training network with EM-fc using (a) $L_{CE}$, (b) $L_T$, (c) $L_J$, and (d) $L_{JMF}$ loss functions.

*B. Impact of Loss Functions, Features and Embedding Modules*

In this section, the impact of loss functions on recognition based on different features is evaluated. The proposed networks are trained on the NTU-PI-v1's training set with different loss functions. Moreover, we also experiment two types of embedding modules, which extract final features used for recognition. As mentioned in the implementation details (see Section III-D), these modules are: 1) EM-fc containing one fully connected layer and outputting 1 x N embedding, and 2) EM-conv containing one convolutional layer and thus retaining the spatial information of the preceding layer. After training, features from palm, little, ring, middle and index fingers, and thumb images are extracted and used for identification on the NTU-PI-v1 test set. Tables I and II show rank-1 and rank-30 identification accuracy for different losses and features using EM-fc and EM-conv, respectively. Figs 5 and 6 show their corresponding CMC curves.

*1) Fully Connected: EM-fc*

For EM-fc, the highest rank-1 accuracies are achieved after training with $L_{JMF}$ loss (see Table I and Fig. 5). The highest one $L_{JMF}$ outperforms the second best $L_J$ by a margin of 12.05% at rank-1 and 12.78% at rank-30, when using all the concatenated features (hand). The palm features are the most discriminative ones. However, adding fingers and thumb features can increase the accuracy significantly and is the highest for $L_{JMF}$ loss, where rank-1 accuracy increases by 11.6% from 45.02% for palm features to 56.63% for hand features. This effect is almost invisible, especially for $L_J$ and $L_T$, where the accuracy increase is less than 1%. In other words, the accuracy of hand features is only less than 1% higher than accuracy of palm features. As mentioned in Section III-C, it could be caused when the palm features dominate over the other features, during training. This is also highlighted by the lowest accuracies of each finger and thumb for these two $L_J$ and $L_T$ losses, which indicate that the network's learning from fingers and thumb is limited.

When finger/s or thumb are of interest, $L_{JMF}$ loss which not only cares about the hand accuracy but also partial accuracies of each finger and thumb (see eq. (3)), offers competitive improvements of ~10-15% margin, comparing to $L_J$, $L_T$ and $L_{CE}$. Note that we are not aware of any available studies of finger or thumb recognition in uncontrolled environment without user cooperation. Therefore, even though this paper does not directly aim at finger and thumb recognition, the results reported here may serve as an indicator or baseline of accuracies that could be expected in such environment and scenarios.

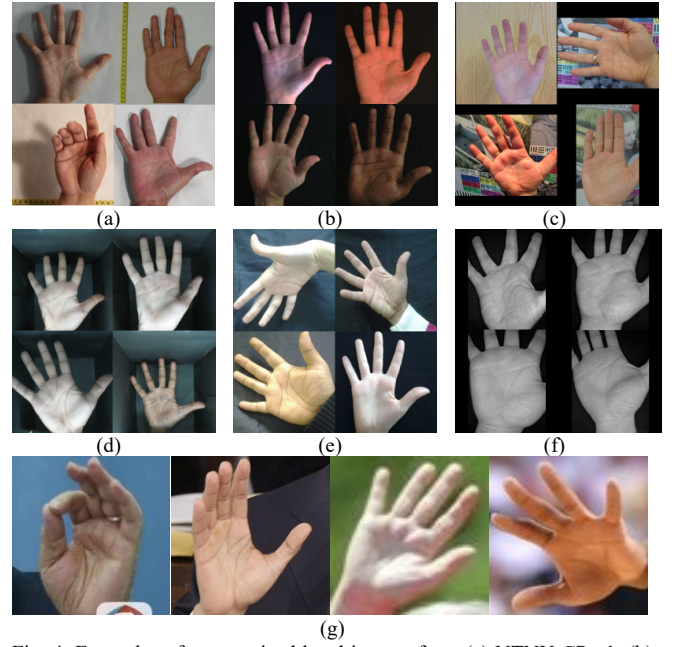

Fig. 4. Examples of square-sized hand images from (a) NTNU-CP-v1, (b) PolyU, (c) NUI_Palm1, (d) IITD, (e) REST and (f) CASIA contactless palmprint databases. (g) Hand images of NTU-PI-v1

*2) Convolutional: EM-conv*

After training with $L_{JMF}$ loss, the EM-conv outperforms the EM-fc at rank-1 by a margin of 5.8%, when using hand features. It achieves 62.42% and 84.76% rank-1 and rank-30 accuracy, respectively. It also outperforms the highest rank-1 and rank-30 accuracy reported in the literature [20] by a margin of 20.5% and 20.03%, respectively, which is close to 50% improvement. As with EM-fc, the palm features achieve the highest rank-1/rank-30 accuracy of 51.44%/75.59% and 52.42%/75.29%, for $L_{JMF}$ and $L_J$, respectively. Nevertheless, it is still significantly lower (margin of ~10%) than when using hand features, which highlights the importance of utilizing full hand image when available in uncontrolled and uncooperative environments, rather than focusing on one part, e.g., palm.

Interestingly, replacing the fully connected layer (EM-fc) with convolutional layer (EM-conv) can reduce the gap in accuracy of hand features among different loss functions. In particular, for EM-conv, the rank-1 accuracy of hand features is 60.91%, 61.68% and 62.42% for $L_T$, $L_J$ and $L_{JMF}$, respectively, whereas for EM-fc, their corresponding accuracies are 43.57%, 44.57% and 56.62%. These results indicate higher robustness and more discriminative power from features with spatial information extracted by the fully convolutional networks.

*C. Evaluation on Publicly Available Contactless Palmprint Databases*

Publicly available contactless palmprint databases were established to study personal recognition likely for commercial or governmental applications such as access control, where user has no physical contact with the acquisition device, (due to hygienic reasons). Typically, in the palmprint literature, networks/algorithms' parameters are trained/tuned and tested on the same database, where images are somewhat similar, making their generalization capabilities limited. NTU-PI-v1 database was originally created to study personal recognition based on hand images for forensic applications. However, the potential and significance of this database to other applications has not been exposed. NTU-PI-v1 contains highly diverse Internet images from 2035 different hands. It is significantly larger in terms of hands than the commonly used IITD and CASIA palmprint datasets that contain images from 460 and 618 hands, respectively. Therefore, it is a reasonable candidate that could support network training and achieve good recognition performance when deployed across multiple databases.

In this section, four networks trained on the NTU-PI-v1 training set are evaluated on hand images from the six representative contactless palmprint databases, namely CASIA, IITD, PolyU, NTU-CP-v1, NUI_Palm1 and REST (see Fig. 4). Networks pretrained using $L_J$ and $L_{JMF}$ are selected because they achieved highest identification accuracies in their corresponding EM-conv and EM-fc variants. To demonstrate network generalization capabilities, no further fine-tuning on images from these contactless palmprint databases is applied. The features $x_{hand}$ are extracted and 1-NN with cosine similarity is used for recognition. In addition to rank-1 accuracies, equal error rates (EER) are also reported in Table III for the settings with EM-fc and EM-conv. EER, which measures the verification performance is used because these contactless palmprint databases are designed for verification scenarios as well, e.g., access control or time and attendance. The rank-1 identification accuracies/EER achieved in this experiment are comparable with the highest/lowest ones reported in the literature [20], [67], [69], [22], [70] on the respective databases, in the same evaluation setting. The highest rank-1 accuracies and the lowest EERs are achieved by EM-conv type, which also is in accord with the results from Section IV-B. It achieves rank-1 accuracies of 97.88%, 100%, 100%, 99.64%, 95.96% and 98.07% on NTU-CP-v1, IITD, PolyU, CASIA, REST and NUI_Palm1, respectively. The lowest reported EER of the proposed networks are 1.54%, 0.59%, 0.45%, 1.34%, 4.94% and 5.81% on NTU-CP-v1, IITD, PolyU, CASIA, REST and NUI_Palm1, respectively. Note that the point of these experimental results is not to outperform all possible results reported in the literature but to show that training the proposed algorithm on NTU-PI-v1 can result in good generalization capabilities. Moreover, it should be noted that in this paper, rather that fine-tuning network to a particular database domain and using pre-processed or pre-selected ROI images, the network trained on NTU-PI-v1 is directly and end-to-end applied to all the hand images (not ROI images) from six different databases, to extract features and perform recognition using 1-NN classifier.

### D. Comparison with Palmprint Recognition Methods on NTU-PI-v1 and analysis of improvements.

This section compares the proposed algorithm using EM-conv with the representative and state-of-the-art palmprint recognition methods on NTU-PI-v1 database. Eight competing methods with top results on NTU-PI-v1 database reported in [20], namely EE-PRnet+PLS, OrdinalCode, DoN, DOC, PalmNet, PCANet, CR-CompCode and DSCN are selected. For fair comparison, in the proposed algorithm, only features $x_{palm}$ are extracted because they are from the palm image and the competing methods are designed for palmprint. 1-NN is used for recognition. The rank-1 and rank-30 identification accuracies are reported in Table IV and the corresponding CMC curves are given in Fig. 7. The proposed algorithm with $L_{JMF}$ and $L_J$ outperforms the second best, namely EE-PRnet, which was also designed for palmprint recognition in uncontrolled and uncooperative environment, with a margin of 9.52% and 10.5% at rank-1, respectively.

TABLE III
IDENTIFICATION ACCURACIES (%) AND EER (%) OF THE FOUR NETWORK PRE-TRAINED ON NTU-PI-V1 AND EVALUATED ON SIX CONTACTLESS PALMPRINT DATABASES.

| | | NTU-CP-v1 | | IITD | | PolyU | | CASIA | | REST | | NUI_Palm1 | |
|---|---|---|---|---|---|---|---|---|---|---|---|---|---|
| Loss | EM- | Rank-1 | EER | Rank-1 | EER | Rank-1 | EER | Rank-1 | EER | Rank-1 | EER | Rank-1 | EER |
| $L_{JMF}$ | conv | 97.46 | 1.71 | **100.00** | 0.66 | **100.00** | 0.56 | **99.64** | 1.43 | **95.96** | **4.48** | **98.07** | 5.84 |
| $L_{JMF}$ | fc | 96.96 | 2.10 | 99.61 | 1.08 | 99.21 | 1.24 | 99.14 | 2.61 | 94.77 | 5.78 | 96.52 | 9.49 |
| $L_J$ | fc | 89.15 | 3.64 | 98.69 | 2.73 | 96.5 | 2.58 | 98.11 | 3.47 | 90.01 | 8.85 | 95.05 | 11.92 |
| $L_J$ | conv | **97.88** | **1.54** | **100.00** | **0.59** | **100.00** | **0.45** | 99.44 | **1.34** | 95.72 | 4.94 | 97.83 | **5.81** |

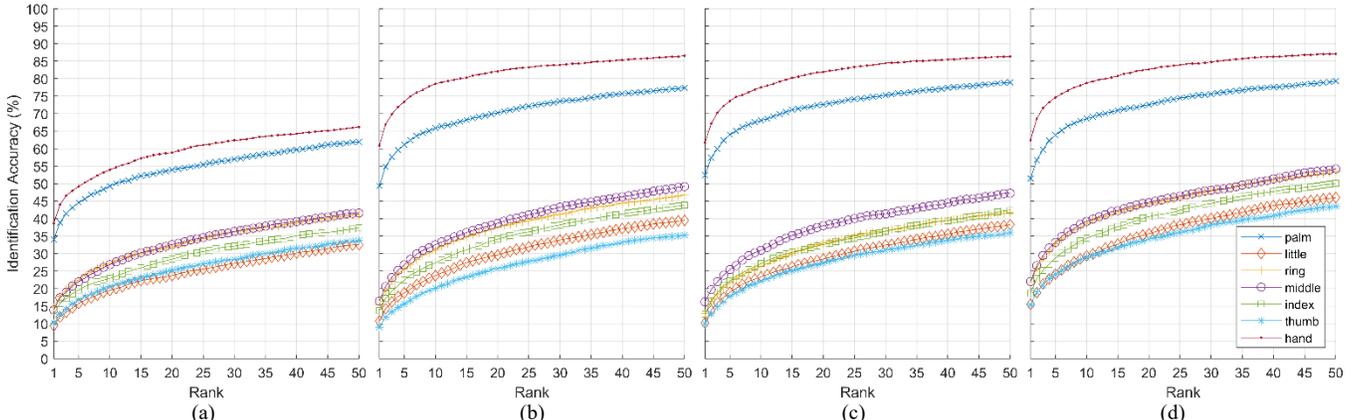

Fig. 6. CMC curves of different features extracted from palm, little, ring, middle, index fingers, thumb and hand, after training network with EM-conv using (a) $L_{CE}$, (b) $L_T$, (c) $L_J$, and (d) $L_{JMF}$ loss functions.

To investigate the improvements of the proposed algorithm over the EE-PRnet [20], which is specifically designed for uncontrolled and uncooperative palmprint images and extensively examined on NTU-PI-v1, two baselines of EE-PRnet are selected for comparison: Baseline-1 with 1-NN and Baseline-2 with PLS. Note that both were trained using cross-entropy loss and Baseline 1 and 2 use 1-NN and PLS for identification, respectively. The improvements in the identification accuracy of the proposed algorithm, considering only palm features $x_{palm}$ and comparing with the two baselines come from three different components: 1) backbone network, 2) training loss functions and classifier and 3) the embedding module EM (penultimate layer). Table V presents the comparison of components and their major contributions to rank-1 accuracy improvements. In the proposed algorithm, the improvement from backbone, comparing to EE-PRnet Baseline-1 using the same loss function and 1-NN classifier is 4.05%. and the further improvement from EM-conv, which contains convolutional layer is 5.03%. (see Table V rows: 1-3). For the setting with fully connected (fc) penultimate layers, comparing the highest rank-1 accuracy reported for EE-PRnet Baseline-2, the major improvements come from loss functions. The experimental results show that the loss functions which contain triplet loss $L_t$, $L_J$ and $L_{JMF}$, can improve the rank-1 accuracy by the margin of 1.3%, 1.86% and 3.1%, respectively, without the need of training a PLS for identification on top (see Table V rows: 4-7). As with the results from Section IV-C-2, the highest rank-1 accuracies are achieved when using convolutional layer (conv) in EM-conv rather than fully connected (fc) one in EM-fc. EM-conv outperforms EM-fc with $L_t$, $L_J$ and $L_{JMF}$, with a margin of 6.15%, 8.84% and 6.42%, respectively (see Table V rows 8-10).

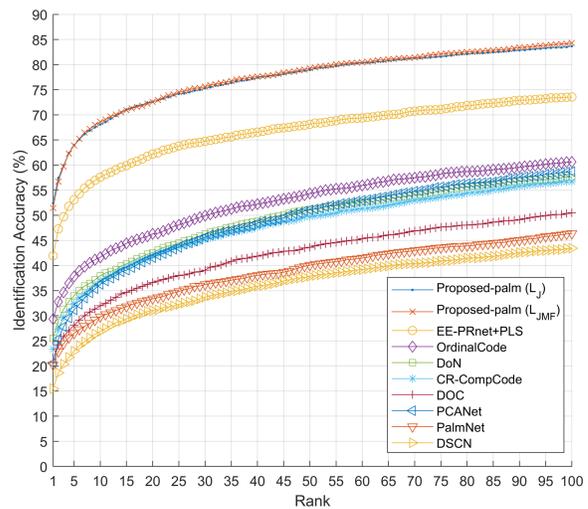

Fig. 7. CMC curves of palmprint recognition methods on NTU-PI-v1 database.

TABLE IV
RANK-1 AND RANK-30 IDENTIFICATION ACCURACIES OF DIFFERENT PALMPRINT RECOGNITION METHODS ON NTU-PI-v1 DATABASE.

| Method | Rank-1 | Rank-30 |
|---|---|---|
| Proposed - palm ($L_{JMF}$) | 51.44 | **75.59** |
| Proposed - palm ($L_J$) | **52.42** | 75.29 |
| EE-PRnet + PLS | 41.92 | 64.73 |
| OrdinalCode | 29.34 | 49.94 |
| DoN | 25.47 | 46.30 |
| DOC | 20.68 | 39.08 |
| PalmNet | 20.21 | 36.21 |
| PCANet | 20.56 | 45.62 |
| CR-CompCode | 23.37 | 45.27 |
| DSCN | 15.47 | 33.70 |

## V. CONCLUSIONS

Using hand-based biometrics for personal identification / verification has been studied by the biometric community for many years. Except methods designed for latent palmprints and latent fingerprints, almost all the existing methods are developed for commercial and government applications, such as access control. Only EE-PRnet [20] is developed for using palmprints collected in uncontrolled and uncooperative environments for forensic investigation. However, its accuracy on NTU-PI-v1 is relatively low, only with rank-1 accuracy of 49.92% and rank-30 accuracy of 64.73%. To achieve higher performance, in this paper, the proposed algorithm consisting of a multiple spatial transformers network, enhanced backbone networks, fully convolution embedding layers and multiple training loss functions to fully utilize the information on fingers, thumb and palm is proposed. Extensive experiments have been conducted. The proposed algorithm achieves rank-1 accuracy of 62.42% and rank-30 accuracy of 84.76% on NTU-PI-v1, which significantly outperforms EE-PRnet and other related state-of-the-art methods designed for contactless palmprint recognition. Its generalization capability is also studied in this paper.

TABLE V
COMPARISON OF RANK-1 ACCURACY IMPROVEMENTS BASED ON DIFFERENT ALGORITHM'S COMPONENTS.

| | Backbone | Classifier | EM | Loss | Rank-1 (%) | Improvement (%.) | Ref. ID | Improvement w.r.t. ref. ID |
|---|---|---|---|---|---|---|---|---|
| EE-PRnet (Baseline 1) | VGG-16 | 1-NN | fc | $L_{CE}$ | 24.91 | | B1 | |
| This paper | ResNet-50 | 1-NN | fc | $L_{CE}$ | 28.96 | +4.05 | fc-CE | B1 |
| This paper | ResNet-50 | 1-NN | conv | $L_{CE}$ | 33.99 | +5.03 | conv-CE | fc-CE |
| EE-PRnet (Baseline 2) | VGG-16 | PLS | fc | $L_{CE}$ | 41.92 | | B2 | |
| This paper | ResNet-50 | 1-NN | fc | $L_T$ | 43.22 | +1.30 | fc-T | B2 |
| | | | | $L_J$ | 43.78 | +1.86 | fc-J | B2 |
| | | | | $L_{JMF}$ | 45.02 | +3.10 | fc-JMF | B2 |
| This paper | ResNet-50 | 1-NN | conv | $L_T$ | 49.37 | +6.15 | conv-T | fc-T |
| | | | | $L_J$ | 52.42 | +8.64 | conv-J | fc-J |
| | | | | $L_{JMF}$ | 51.44 | +6.42 | conv-JMF | fc-JMF |

ACKNOWLEDGMENT

The work is partially supported by NTU Internal Funding - Accelerating Creativity and Excellence (NTU–ACE2020-03).